\documentclass{article}

\usepackage{PRIMEarxiv}

\usepackage[utf8]{inputenc} 
\usepackage[T1]{fontenc}    
\usepackage{hyperref}       
\usepackage{url}            
\usepackage{booktabs}       
\usepackage{amsfonts}       
\usepackage{nicefrac}       
\usepackage{microtype}      
\usepackage{lipsum}
\usepackage{fancyhdr}       
\usepackage{graphicx}       
\graphicspath{{media/}}     
\usepackage{amsmath,amssymb,amsfonts}
	
	\usepackage{graphicx}
	\usepackage{textcomp}
        \usepackage{breqn}
	\usepackage{enumitem}
        \usepackage{tikz}
        \usetikzlibrary{arrows.meta, positioning, fit, backgrounds}

	\usepackage{amsthm}
	\newtheorem{theorem}{Theorem}

	\usepackage{graphicx}
        \usepackage{subcaption}
        \usepackage{algorithm}
        \usepackage{algpseudocode}
\newcommand{\BEAS}{\begin{eqnarray*}}
	\newcommand{\EEAS}{\end{eqnarray*}}
\newcommand{\BEA}{\begin{eqnarray}}
	\newcommand{\EEA}{\end{eqnarray}}
\newcommand{\BEQ}{\begin{equation}}
	\newcommand{\EEQ}{\end{equation}}
\newcommand{\BIT}{\begin{itemize}}
	\newcommand{\EIT}{\end{itemize}}

\newcommand{\ie}{{\it i.e.}}

\newcommand{\diag}{\mathop{\rm diag}}

\newcommand{\Prob}{\mathop{\mathbb{P}}}

\newcommand{\argmin}{\mathop{\rm argmin}}

\newcounter{exno}
\newlength{\exlabelwidth}

{\clearpage\section*{Exercises}%
	\addcontentsline{toc}{section}{Exercises}%
	\markright{Exercises}%
	\begin{small}\addtolength{\baselineskip}{-1.0pt}%
		\begin{list}{\bfseries\textsf{\thechapter.\arabic{exno}}}%
			{\usecounter{exno}%
				\settowidth{\exlabelwidth}{\bfseries\textsf{\thechapter.99}}%
				\setlength{\labelwidth}{\exlabelwidth}%
				\setlength{\labelsep}{.5em}%
				\setlength{\leftmargin}{0mm}%
				\setlength{\rightmargin}{0mm}}} %
		{\end{list}\clearpage%
	\end{small}\addtolength{\baselineskip}{1.0pt}}   

\makeatletter
\long\def\@makecaption#1#2{
	\vskip 9pt 
	\begin{small}
		\setbox\@tempboxa\hbox{{\sffamily\bfseries #1} #2}
		\ifdim \wd\@tempboxa > 0.85\textwidth
		\begin{center}
			\begin{minipage}[t]{0.85\textwidth}
				\addtolength{\baselineskip}{-0.95pt}
				{\sffamily\bfseries #1} #2 \par
				\addtolength{\baselineskip}{0.95pt}
			\end{minipage}
		\end{center}
		\else 
		\hbox to\hsize{\hfil\box\@tempboxa\hfil}  
		\fi
	\end{small}\par
}
\makeatother

\newcounter{oursection}

\newcounter{lecture}

\newcommand{\calP}{{\mathcal P}}

\newcommand{\calB}{{\mathcal B}}
\newcommand{\calL}{{\mathcal L}}

\newcommand{\N}{{\mathbb{N}}}
\newcommand{\reals}{{\mathbb{R}}}
\newcommand{\stilde}{{\tilde{s}}}
\newcommand{\btilde}{{\tilde{b}}}

\newcommand{\Qhat}{{\hat{Q}}}
\newcommand{\Chat}{{\hat{C}}}

\newcommand{\Cbar}{{\bar{C}}}
\newcommand{\sbar}{{\bar{s}}}
\newcommand{\pibar}{{\bar{\pi}}}

\title{Markov Decision Processes\\ with Noisy State Observation
}

\author{
  Amirhossein Afsharrad, Sanjay Lall \\
  Department of Electrical Engineering \\
  Stanford University \\
  Stanford, CA\\
  \texttt{\{afsharrad, lall\}@stanford.edu} 
}

\begin{document}
\maketitle
\begin{abstract}
This paper addresses the challenge of a particular class of noisy state observations in Markov Decision Processes (MDPs), a common issue in various real-world applications. We focus on modeling this uncertainty through a confusion matrix that captures the probabilities of misidentifying the true state. Our primary goal is to estimate the inherent measurement noise, and to this end, we propose two novel algorithmic approaches. The first, the method of second-order repetitive actions, is designed for efficient noise estimation within a finite time window, providing identifiable conditions for system analysis. The second approach comprises a family of Bayesian algorithms, which we thoroughly analyze and compare in terms of performance and limitations. We substantiate our theoretical findings with simulations, demonstrating the effectiveness of our methods in different scenarios, particularly highlighting their behavior in environments with varying stationary distributions. Our work advances the understanding of reinforcement learning in noisy environments, offering robust techniques for more accurate state estimation in MDPs.
\end{abstract}


\section{Introduction}
In the landscape of artificial intelligence, Markov Decision Processes (MDPs) are pivotal in reinforcement learning, providing a structured approach to tackle sequential decision-making problems under uncertainty \cite{van2012reinforcement, sigaud2013markov}. Their applicability spans a diverse range of fields, including robotics \cite{kurniawati2022partially, zhang2021reinforcement}, financial decision-making \cite{bauerle2011markov, petrik2012approximate}, autonomous navigation \cite{kinjo2013evaluation, zhang2019decision}, healthcare management \cite{bennett2013artificial, gocgun2011markov}, supply chain optimization \cite{kemmer2018reinforcement, ahiska2013markov}, and energy systems management \cite{hansen2016partially, nanduri2013competitive}. MDPs enable the optimization of actions based on probabilistic state transitions, which is fundamental in these domains. However, a critical challenge emerges when these theoretical models are applied to real-world scenarios characterized by noisy observations and measurements. This is a common occurrence in environments ranging from sensor data in autonomous vehicles \cite{jo2015precise} to fluctuating markets in finance \cite{li2019deep}, from patient data in healthcare \cite{yu2021reinforcement} to demand variability in supply chains \cite{mao2016resource}, and from renewable energy predictions to grid stability in energy systems \cite{perera2021applications}. Such inherent uncertainties in observations necessitate sophisticated methodologies for accurate state estimation, which is essential for making reliable and informed decisions. Our work addresses this challenge, proposing advanced techniques to enhance the robustness and effectiveness of MDP-based reinforcement learning models in the face of noise in state observations.

In this work, we address a particular type of noisy observations by modeling the uncertainty through a confusion matrix, which represents the probabilities of observing states incorrectly. Our approach adopts a system identification perspective, aiming to estimate the inherent measurement noise characterized by this confusion matrix. The primary contribution of this paper lies in proposing two categories of algorithms designed to estimate these noise components, accompanied by a set of conditions and performance guarantees. Initially, we introduce an innovative method, \textit{the method of second-order repetitive actions}. This approach efficiently estimates noise components within a finite time window, and we specify the conditions for the reliable identification of the system's noise characteristics. Subsequently, we introduce a family of Bayesian algorithms, providing a comparative analysis of their performance and exploring their potential limitations. Through these contributions, we aim to enhance the robustness and accuracy of MDPs in environments with observational noise.

    \section{Problem Formulation}
    We consider a standard reinforcement learning problem defined by a Markov Decision Process (MDP). The MDP is characterized by a finite set of states, denoted as $S = \{1,\cdots, n\}$ where $n$ is the number of states, and a finite set of actions, represented as $A = \{1,\cdots, m\}$ with $m$ being the number of actions. Denoting $\calP_n$ as the set of all $n\times n$ transition probability matrices, the transition probability function $P:A\to \calP_n$ maps actions to $n\times n$ transition probability matrices, defined as $P(a)_{ij}=\Prob\left[s_{t+1}=j|s_t=i, a_t=a\right]$, where $t\in\N$ is the time-step, and $a_t$ is the action at time $t$. The reward function $R:S\times A\times \reals \to \reals$ determines the immediate reward, $r_t=R(s_t,a_t, \eta_t)$, with $r_t$ as the observed reward at time $t$ and \(\eta_t\) as a noise variable, incorporated in \(R\) to capture stochastic dependencies of the reward on state-action pairs.

    This study focuses on the challenge of noisy state observation in MDPs. Considering the finite nature of the state space, we explore state observation noise by defining the probability of observing a state $\stilde\in S$ different from the actual state $s\in S$. In this context, $\stilde$ and $s$ represent the \textit{observed state} and the \textit{true state}, respectively. This noise is characterized by the state observation confusion matrix $C\in\reals^{n\times n}$, which specifies the probabilities of observing $\stilde$ when the actual state is $s$, expressed as $C_{ij} = \Prob\left[\stilde_t=j|s_t=i\right]$. This matrix is assumed to be time-independent, indicating consistent observation noise characteristics across states. While $S$, $A$, and $P$ are known, the reward function $R$ and the confusion matrix $C$ are unknown. Our objective is to estimate $C$, and given our focus on estimating $C$ in settings with absent or sparse rewards, we do not assume a specific reward model, ensuring that our algorithms are not dependent on reward dynamics.

    \section{Main Results}
    \subsection{The Method of Second-Order Repetitive Actions}\label{sec:rep_act}
        \begin{theorem}\label{thm:quad}
            Consider an MDP with noisy observations and observation confusion matrix $ C $. Define $ Q^{a}_t\in\reals^{n\times n} $ as the observed state transition matrix at time $ t $, where 
            \[Q^{a}_{t,ij} = \Prob\left[\stilde_{t+1}=j|\stilde_t=i, a_t=a\right].\]
            Let $ \pi_t\in\reals^n $ represent the probability distribution over the true state at time $ t $, expressed as 
            \[\pi_t = \Prob\left[s_t=i\right].\]
            Then, the following relationship holds:
            \begin{equation}\label{eq:quad_eq}
                \diag(C^\top  \pi_t)Q^{a}_t=C^\top \diag(\pi_t)P(a)C.
            \end{equation}
        \end{theorem}
        Proof provided in Section \ref{sec:quad_thm_proof}.
        
        Our main goal is to deduce $ C $ from $ Q^{a}_t $. An empirical estimate $ \hat{Q}^{a}_t $ enables us to solve \eqref{eq:quad_eq} for $ C $. The primary challenge is the time-dependent nature of $ Q^{a}_t $, as single observations per time-step are insufficient for reliable estimation of $ \Qhat^{a}_t $.

        To address this, we note that the time dependence in $ Q^{a}_t $ is solely through $ \pi_t $. Stabilizing $ \pi_t $ over a period allows $ Q^{a}_t $ to become time-independent, enabling us to empirically estimate \(Q^a_t\) with $ \Qhat^a $ and then solve \eqref{eq:quad_eq} for $ C $. We achieve this by repeatedly selecting an action $ a\in A $ for a duration, leading $\pi_t$ to converge to the stationary distribution $ \pi^a $ of $ P(a) $, under the assumption that the Markov process for \(P(a)\) is irreducible and aperiodic. The observed state transitions then adhere to a constant matrix $ Q^a $, defined by 
        \begin{equation}\label{eq:quad_eq_stationary}
            \diag(C^\top  \pi^a)Q^a=C^\top \diag(\pi^a)P(a)C.
        \end{equation}
        This allows us to empirically determine $ \Qhat^a $ and estimate $ C $ by minimizing the loss function $ \calL_a $ defined as
        \begin{equation}\label{eq:loss_func}
            \calL_a(C) = \left\|C^\top \diag(\pi^a)P(a)C-\diag\left(C^\top \pi^a\right)\Qhat^a\right\|^2.
        \end{equation}
        However, this approach can lead to non-unique solutions for $ C $, as demonstrated with an example for $n=2$ later. Different $ C $ matrices can produce the same $ Q^a $, making it impossible to uniquely determine $ C $ from \eqref{eq:loss_func}, even with the perfect esitmate $\Qhat^a = Q^a$.

        To overcome this, we apply the method across various actions. While each action might yield multiple solutions, the common solution across different actions may uniquely identify the actual $ C $. We select a subset $ A_0 $ of $ A $ with $ |A_0|=N $, and for each action in $ A_0 $, repeated for a duration $ T $, we estimate $ \Qhat^a $ and solve for $ C $ using:
        \begin{equation}\label{eq:loss_func_n}
            \Chat=\argmin_C \frac{1}{N} \sum_{a\in A_0}^{}\left\| C^\top \diag(\pi^a)P(a)C-\diag\left(C^\top \pi^a\right)\Qhat^a \right\|^2.
        \end{equation}
        However, even with multiple actions from $ A_0 $, \eqref{eq:loss_func_n} may have multiple solutions. As an example, consider two actions \(a, a'\) with the following probabilities
        \[
		P(a)=\begin{bmatrix}
			0 & 1 \\ 1 & 0
		\end{bmatrix}, \quad 
		Q^a=\begin{bmatrix}
			0.45 & 0.55 \\ 0.825 & 0.175
		\end{bmatrix}, \quad 
            P(a')=\begin{bmatrix}
			0.3 & 0.7 \\ 0.7 & 0.3
		\end{bmatrix}, \quad 
		Q^{a'}=\begin{bmatrix}
			0.54 & 0.46 \\ 0.69 & 0.31
		\end{bmatrix}.
		\]
        Solving \eqref{eq:quad_eq_stationary} for both actions, we find two possible solutions for $ C $ given by
        \[
		C_1=\begin{bmatrix}
			0.9 & 0.1 \\ 0.3 & 0.7
		\end{bmatrix}, \quad 
		C_2=\begin{bmatrix}
			0.3 & 0.7 \\ 0.9 & 0.1
		\end{bmatrix},
	\]
                suggesting that both \(C_1\) and \(C_2\) are solutions of \eqref{eq:loss_func_n}, illustrating the non-uniqueness issue. This example illustrates that multiple actions do not guarantee a unique solution for $ C $. Theorem \ref{thm:2_unique} addresses this issue by providing a necessary and sufficient condition for uniqueness in the $n=2$ case.
        \begin{theorem}\label{thm:2_unique}
            For \(n=2\), a unique solution for 
            \begin{equation} \label{quad_eq_thm}
                C^\top \diag(\pi^a)P(a)C=\diag\left(C^\top \pi^a\right)Q^a, \quad a\in \{a_1,a_2\}
            \end{equation}
            exists iff the stationary distributions of $P(a_1)$ and $P(a_2)$ are distinct, \ie, $\pi^{a_1}\neq\pi^{a_2}$.
        \end{theorem}
        Proof in Section \ref{sec:2_unique_thm_proof}.	Theorem \ref{thm:n_unique} generalizes this condition for $n\in\N$:
        \begin{theorem}\label{thm:n_unique}
            Let $B$ be any non-empty strict subset of $\N_n=\{1,2,\cdots, n\}$, and define $\calB=\{B\subset \N_n|B\neq \emptyset, \N_n\}$ as the set of all such subsets.
			Then, if \(C\) is symmetric, the set of equations 
			\begin{equation}\label{n_eq}
			    C^\top \diag(\pi^a)P(a)C=\diag\left(C^\top \pi^a\right)Q^a, \quad a\in A_0
			\end{equation}
			has a unique solution if for every $ B\in\calB $, there exist two actions $ a,a'\in A_0 $ such that 
			\begin{equation}\label{eq:n_unique_condition}
			    \sum_{i\in B}^{}\pi^a_i\neq \sum_{i\in B}^{}\pi^{a'}_i.
			\end{equation}
        \end{theorem}
        Proof in Section \ref{sec:n_unique_thm_proof}. Note that Theorem \ref{thm:n_unique}'s uniqueness condition, while sufficient, is not strictly necessary and can be relaxed in various ways. The assumption of a symmetric \(C\) simplifies the proof but is not essential.

        To conclude, the second-order repetitive actions algorithm, involving repetitive actions and recording the last two observed states, estimates \(Q^a\) and \(C\). It is called a second-order algorithm as it involves recording only the last two observations and results in a second order equation in \(C\). Theorems \ref{thm:2_unique} and \ref{thm:n_unique} ensure solution convergence to the true \(C\). The next section introduces other methods for estimating the confusion matrix.

        \subsection{The Bayesian Method}\label{sec:bayes}
    In this section, we apply a Bayesian approach to estimate the confusion matrix $C$. This method involves maintaining two probability distributions: one over the unknown parameters characterized by \(C\), and the other over the current state of the system given past observations. We introduce two variants of the Bayesian method and analyze their performances in Section \ref{sec:sim}.

    \subsubsection{The First-Order Bayesian Algorithm}\label{sec:bayes1}
    The belief state $b_t\in\reals^n$ is defined as
    \begin{equation}\label{eq:belief1}
        b_{t,i} = \Prob\left[s_t=i|\stilde_t,a_{t-1},\cdots, a_1,\stilde_0\right],
    \end{equation}
    and the probability density function $p_t:\reals^{n\times n}\to \reals^+$ gives the probability density at $C$ at time $t$ conditioned on past observations and actions, \ie,
    \begin{equation}\label{eq:c_prob}
        p_t(C_0) = \Prob\left[C=C_0|\stilde_t,a_{t-1},\cdots, a_1,\stilde_0\right].
    \end{equation}

    \begin{theorem}\label{thm:bayes1}
        Given a history of observations and actions, the belief state $b_t$, and the probability density function $p_t$ at $C$, the update rules are as follows:
        \begin{enumerate}[label=(\alph*)]
            \item The update rule for the probability density function of $C$:
            \begin{align}
                p_{t+1}(C)&= \frac{b_t^\top P C_{\stilde_{t+1}}}
                {\int_{C'} b_t^\top P C'_{\stilde_{t+1}}p_t(C')\mathrm{d}C'} p_t(C),
            \end{align}
            where $C_{\stilde_{t+1}}$ is the $\stilde_{t+1}$th column of $C$.
            \item The belief state update rule:
            \begin{equation}
                b_{t+1,i} = \int_C\frac{C_{i, \stilde_{t+1}}\sum_{j} P(a_t)_{ji}b_{t,j}}{\sum_{k}C_{k, \stilde_{t+1}}\sum_{j} P(a_t)_{jk}b_{t,j}}p_{t+1}(C)
                \mathrm{d}C
            \end{equation}
        \end{enumerate}
    \end{theorem}

    \subsubsection{The Second-Order Bayesian Algorithm}\label{sec:bayes2}
    The second-order Bayesian algorithm incorporates the last two observed states into the probability computation. 
    It requires a modified belief state, denoted by \(\btilde_t\), defined as
    \[\btilde_{t,i} = \Prob\left[s_t=i|\stilde_{t-1},a_{t-2},\cdots, a_1,\stilde_0\right].\]
    \begin{theorem}\label{thm:bayes2}
        Given a history of observations and actions, the current belief state $\btilde_t$, and the probability density function $p_t$ at $C$, the update rules are:
        \begin{enumerate}[label=(\alph*)]
            \item The update rule for the probability density function of $C$:
            \begin{align}
                Q(C) &= \diag\left(C^\top \btilde_t\right)^{-1}C^\top  \diag(\btilde_t) P(a_t) C \\
                p_{t+1}(C)&= \frac{Q(C)_{\stilde_t, \stilde_{t+1}} }{\int_{C'} Q(C')_{\stilde_t, \stilde_{t+1}}p_t(C')\mathrm{d}C'} p_t(C).
            \end{align}
            \item The belief state update rule:
            \begin{equation}
                \btilde_{t+1,i} = \int_C\frac{\sum_j P(a_t)_{ji}C_{j, \stilde_{t}}\btilde_{t,j}}{\sum_j C_{j, \stilde_t}\btilde_{t,j}}p_{t+1}(C)
                \mathrm{d}C
            \end{equation}
        \end{enumerate}
    \end{theorem}
    The proof of this theorem, similar to that of Theorems \ref{thm:quad} and \ref{thm:bayes1}, is omitted for brevity. Theorems \ref{thm:bayes1} and \ref{thm:bayes2} provide methods for maintaining and updating probability distributions over \(C\) and the true state. Their performance will be compared in Section \ref{sec:sim}. The Bayesian approach offers flexibility in action choice, unlike the second-order repetitive actions method in Section \ref{sec:rep_act} that requires repeating the same action. However, maintaining a probability distribution over \(n\times n\) matrices may be computationally intensive for large \(n\).

    \section{Proofs}\label{sec:proofs}
    \subsection{Proof of Theorem \ref{thm:quad}}\label{sec:quad_thm_proof}
        We derive the relationship between the matrices by conditioning over the states \(s_t\) and \(s_{t+1}\). For an arbitrary element of \(Q^{a}_t\), we have
		\begin{align*}
			Q^{a}_{t,ij} &= \Prob\left[\stilde_{t+1}=j|\stilde_t=i, a_t=a\right]\\
			&= \sum_{k=1}^{n} \Prob\left[s_{t+1}=k|\stilde_t=i, a_t=a\right]\underbrace{\Prob\left[\stilde_{t+1}=j|\stilde_t=i, s_{t+1}=k, a_t=a\right]}_{C_{kj}}\\
			&= \sum_{k=1}^{n}C_{kj}\sum_{\ell=1}^{n} \Prob\left[s_t=\ell|\stilde_t=i, a_t=a\right]\underbrace{\Prob\left[s_{t+1}=k|\stilde_t=i, s_t=\ell, a_t=a\right]}_{P(a)_{\ell k}}\\
			&= \sum_{k=1}^{n}C_{kj}\sum_{\ell=1}^{n} P(a)_{\ell k}\frac{\Prob[s_t=\ell|a_t=a]\Prob\left[\stilde_t=i|s_t=\ell, a_t=a\right]}{\sum_{\ell'=1}^{n}\Prob[s_t=\ell'|a_t=a]{\Prob\left[\stilde_t=i|s_t=\ell', a_t=a\right]}}\\
			&= \sum_{k=1}^{n}C_{kj}\sum_{\ell=1}^{n} P(a)_{\ell k}\frac{\Prob[s_t=\ell]\Prob\left[\stilde_t=i|s_t=\ell\right]}{\sum_{\ell'=1}^{n}\Prob[s_t=\ell']{\Prob\left[\stilde_t=i|s_t=\ell'\right]}}\\
                &= \frac{1}{\sum_{\ell'=1}^{n}\pi_{t,\ell'}C_{\ell'i}}\sum_{k=1}^{n}\sum_{\ell=1}^{n} C_{kj}P(a)_{\ell k}\pi_{t,\ell}C_{\ell i}.\\
		\end{align*}
                Multiplying both sides by \(\sum_{\ell'=1}^{n}\pi_{t,\ell'}C_{\ell'i}\) and expressing in matrix form, we obtain \(\diag\left(C^\top \pi_t\right)Q^{a}_t = C^\top \diag(\pi_t)P(a)C\), concluding the proof.

    \subsection{Proof of Theorem \ref{thm:2_unique}}\label{sec:2_unique_thm_proof}
    Consider 
    \[
    C=\begin{bmatrix}
        c& 1-c\\ 
        d& 1-d
    \end{bmatrix}, \quad
    P(a)=\begin{bmatrix}
        p& 1-p\\ 
        p'& 1-p'
    \end{bmatrix}, \quad
    \pi^a=\begin{bmatrix}
        \alpha\\ 
        1-\alpha
    \end{bmatrix},
    \]
    where \(C\) is the confusion matrix, \(P(a)\) is the state transition probability matrix for action \(a\), and \(\pi^a\) is its stationary distribution. Given \(P(a)^\top \pi^a = \pi^a\), we can solve for \(p'\) in terms of \(\alpha\) and \(p\), and rewrite \(P(a)\) as
    \[P(a)=\begin{bmatrix}
        p& 1-p\\ 
        \frac{\alpha}{1-\alpha}(1-p)& 1-\frac{\alpha}{1-\alpha}(1-p)
    \end{bmatrix}.\]
    We have \(Q^a=\diag(C^\top\pi^a)^{-1}C^\top\diag(\pi^a)P(a)C\). We are interested in solutions to
    \begin{equation}\label{quad_eq}
        X^\top \diag(\pi^a)P(a)X=\diag\left(X^\top \pi^a\right)Q^a,
    \end{equation}
    where \(X\) is the unknown matrix. Representing \(X\) as 
    \(\begin{bmatrix}
        x & 1-x \\ 
        y & 1-y 
    \end{bmatrix}\), 
    \eqref{quad_eq} becomes a system of two quadratic equations in \(x,y\). Geometrically, this equates to finding the intersection points of two ellipsoids in \(\reals^2\), each intersection corresponding to a solution \(X\). Let these ellipsoids be represented as 
    \begin{equation}\label{eq:2_solutions}
        \begin{cases}
        A_{xx}x^2+A_{yy}y^2+A_{xy}xy+A_xx+A_yy+A_0=0 \\
        B_{xx}x^2+B_{yy}y^2+B_{xy}xy+B_xx+B_yy+B_0=0,
    \end{cases}
    \end{equation}
    with coefficients determined in terms of \(\alpha, p, c, d\) from \eqref{quad_eq}. Solving the first ellipsoid equation for \(x^2\) and substituting it into the second results in an equation of the form \(C_xx+C_yy+C_0=0\). This simplifies the problem to finding intersections of a line and an ellipsoid, implying at most two solutions. The intersection points are 
    \begin{equation}\label{quad_sols}
        \begin{cases}
        (x_1, y_1) = (c, d) \\
        (x_2,y_2)=(2d - c + 2\alpha c - 2\alpha d, d + 2\alpha c - 2\alpha d).
    \end{cases}
    \end{equation}
    These solutions are independent of \(p\), depending only on \(\pi^a\). Thus, if \(P(a_1)\) and \(P(a_2)\) have the same stationary distributions, both corresponding confusion matrices from \eqref{quad_sols} satisfy \eqref{quad_eq_thm}, indicating non-uniqueness. For different stationary distributions characterized by \(\alpha\) and \(\alpha'\), non-uniqueness implies 
    \[d + 2\alpha c - 2\alpha d=d + 2\alpha' c - 2\alpha' d\Rightarrow \alpha=\alpha' \text{ or } c=d.\]
    Since \(\alpha\neq\alpha'\) by assumption, \(c=d\) must hold. However, if \(c=d\), then \((x_1,y_1)=(x_2,y_2)=(c,d)\), leading to unique solutions. Thus, different stationary distributions of \(P(a_1)\) and \(P(a_2)\) are necessary and sufficient for the uniqueness of the solution of \eqref{quad_eq_thm}, concluding the proof.

    \subsection{Proof of Theorem \ref{thm:n_unique}}\label{sec:n_unique_thm_proof}
    To establish a sufficient condition for the uniqueness of the solution of \eqref{n_eq}, we leverage the result of Theorem \ref{thm:2_unique}. We consider an MDP with \(n\) states and partition these states into two \textit{superstates}, \(\sbar_1\) and \(\sbar_2\). Without loss of generality, let \(\sbar_1=\{s_1,\cdots, s_k\}\) and \(\sbar_2=\{s_{k+1},\cdots, s_n\}\). The stationary distribution and confusion matrix for this partitioned MDP are given by:
    \begin{equation}
        \pibar=\begin{bmatrix}
            \pi_1+\cdots+\pi_k\\
            \pi_{k+1}+\cdots+\pi_n
        \end{bmatrix}, \quad 
        \Cbar = \begin{bmatrix}
            \Cbar_{11} & 1 - \Cbar_{11} \\
            1-\Cbar_{22} & \Cbar_{22}
        \end{bmatrix},
    \end{equation}
    where
    \begin{equation}\label{eq:cbar}
        \Cbar_{11}=\sum_{i=1}^k\frac{\pi_i\sum_{j=1}^k C_{ij}}{\sum_{j=1}^k \pi_j}, \quad \Cbar_{22}=\sum_{i=k+1}^n\frac{\pi_i\sum_{j=k+1}^n C_{ij}}{\sum_{j=k+1}^n \pi_j}.
    \end{equation}
    Examples of such partitioning are shown in Figures \ref{fig:1_3_partition} and \ref{fig:2_2_partition}, representing the state observation confusion.

    \begin{figure}[h]
          \centering
          \begin{minipage}{.49\textwidth}
            \centering
            \begin{tikzpicture}[->,>=Stealth,node distance=2cm, thick, main node/.style={circle, draw, font=\sffamily\Large\bfseries}]

              \node[main node] (1) at (0,1.8) {$s_1$}; 
              \node[main node] (2) at (1.5,0) {$s_2$}; 
              \node[main node] (3) at (0,-1) {$s_3$}; 
              \node[main node] (4) at (-1.5,0) {$s_4$}; 
            
              \foreach \from/\to in {1/2,2/3,3/4,4/1,1/3,3/1,2/4,4/2,1/4,4/1,2/1,1/2,3/2,2/3}
                    \draw (\from) to[] (\to);
            
                \path (1) edge [loop, out=120, in=150, looseness=5] node {} (1);
                \path (2) edge [loop below] node {} ();
                \path (3) edge [loop right] node {} ();
                \path (4) edge [loop below] node {} ();

              \begin{pgfonlayer}{background}
                \node[rounded corners, draw=black!50, dashed, fill=blue!20, fit=(1), inner sep=8pt] (supernode1) {};
                \node[rounded corners, draw=black!50, dashed, fill=blue!20, fit= (2)(3) (4), inner sep=8pt] (supernode2) {};
              \end{pgfonlayer}
            
              \path (supernode2) edge [blue,->, very thick, out=150, in=180, looseness=1] node[above left] {$1-\Cbar_{22}$} (supernode1);
              \path (supernode1) edge [blue,->, very thick, out=0, in=30, looseness=1] node[above right] {$1-\Cbar_{11}$} (supernode2);
            
              \path[blue, very thick] (supernode1) edge [loop above] node[above] {$\Cbar_{11}=C_{11}$} ();
              \path (supernode2) edge [blue,->, very thick, out=280, in=260, looseness=5] node[below] {$\Cbar_{11}=\frac{\pi_2(C_{22}+C_{23}+C_{24})}{\pi_2+\pi_3+\pi_4}+\cdots+\frac{\pi_4(C_{42}+C_{43}+C_{44})}{\pi_2+\pi_3+\pi_4}$} (supernode2);
            \end{tikzpicture}
            \caption{Partitioning with \(\sbar_1=\{s_1\}\)}
            \label{fig:1_3_partition}
          \end{minipage}
          \begin{minipage}{.49\textwidth}
            \centering
            \begin{tikzpicture}[->,>=Stealth,node distance=2cm, thick, main node/.style={circle, draw, font=\sffamily\Large\bfseries}]

              \node[main node] (1) at (0,0) {$s_1$}; 
              \node[main node] (2) at (2.25,0) {$s_2$}; 
              \node[main node] (3) at (0,-2.25) {$s_3$}; 
              \node[main node] (4) at (2.25,-2.25) {$s_4$}; 
            
              \foreach \from/\to in {1/2,2/3,3/4,4/1,1/3,3/1,2/4,4/2,1/4,4/1,2/1,1/2,3/2,2/3}
                    \draw (\from) to[] (\to);
            
                \path (1) edge [loop above] node {} ();
                \path (2) edge [loop above] node {} ();
                \path (3) edge [loop below] node {} ();
                \path (4) edge [loop below] node {} ();

              \begin{pgfonlayer}{background}
                \node[rounded corners, draw=black!50, dashed, fill=blue!20, fit=(1) (2), inner sep=8pt] (supernode1) {};
                \node[rounded corners, draw=black!50, dashed, fill=blue!20, fit=(3) (4), inner sep=8pt] (supernode2) {};
              \end{pgfonlayer}
            
              \draw[blue,->, very thick] (supernode1.east) to[bend left] node[right] {$1-\Cbar_{11}$} (supernode2.east);
              \draw[blue,->, very thick] (supernode2.west) to[bend left] node[left] {$1-\Cbar_{22}$}(supernode1.west);
            
              \path[blue, very thick] (supernode1) edge [loop above] node[above] {$\Cbar_{11}=\frac{\pi_1(C_{11}+C_{12})}{\pi_1+\pi_2}+\frac{\pi_2(C_{21}+C_{22})}{\pi_1+\pi_2}$} ();
              \path[blue, very thick] (supernode2) edge [loop below] node[below] {$\Cbar_{22}=\frac{\pi_3(C_{33}+C_{34})}{\pi_3+\pi_4}+\frac{\pi_4(C_{43}+C_{44})}{\pi_3+\pi_4}$} ();
            \end{tikzpicture}
            \caption{Partitioning with \(\sbar_1=\{s_1,s_2\}\)}
            \label{fig:2_2_partition}
          \end{minipage}%
        \end{figure}

    The assumption in \eqref{eq:n_unique_condition} ensures that for any partitioning, the stationary distributions of the two-state MDP are different for some actions \(a, a'\). This, as a result of Theorem \ref{thm:2_unique}, implies that \(\Cbar_{11}\) and \(\Cbar_{22}\) can be uniquely determined for any partitioning. To uniquely determine \(C\), we first find all diagonal elements \(C_{ii}\) using the partitioning \(\sbar_1=\{s_i\}, \sbar_2=\{s_1,\cdots, s_n\}\setminus\{s_i\}\). To find an arbitrary \(C_{ij}\), we use the partitioning \(\sbar_1=\{s_i, s_j\}, \sbar_2=\{s_1,\cdots, s_n\}\setminus\{s_i, s_j\}\). With \(\Cbar_{11}\) being a linear function of \(C_{ii}, C_{ij}, C_{ji}, C_{jj}\), and knowing \(C_{ii}, C_{jj}\) and the symmetry of \(C\) (\ie, \(C_{ij} = C_{ji}\)), we can solve for \(C_{ij}\). As this applies to any element of \(C\), the entire matrix \(C\) can be uniquely determined. This concludes the proof.

        \subsection{Proof of Theorem \ref{thm:bayes1}}
        The derivation of the two update rules for \(p_t\) and \(b_t\) are given below. 
        \begin{align*}
            p_{t+1}(C) &= \Prob\left[C|\stilde_{t+1},a_{t},\cdots, a_1,\stilde_0\right] \\
            &= \frac{\Prob\left[\stilde_{t+1}|C, a_t, \stilde_t, \cdots, \stilde_0\right]\Prob\left[C|\stilde_t, a_{t-1},\cdots, \stilde_0\right]}{\Prob\left[\stilde_{t+1}, a_t, \stilde_t, \cdots, \stilde_0\right]}\\
            &= \frac{\sum_{i}\Prob\left[\stilde_{t+1}|s_{t+1}=i, C, a_t, \stilde_t, \cdots, \stilde_0\right]\Prob\left[s_{t+1}=i|C,a_t, \stilde_t, \cdots, \stilde_0 \right]p_{t}(C)}{\Prob\left[\stilde_{t+1}, a_t, \stilde_t, \cdots, \stilde_0\right]}\\
            &= \frac{\sum_{i}C_{i, \stilde_{t+1}}\sum_{j}\Prob\left[s_{t+1}=i|s_t=j,C,a_t, \stilde_t, \cdots, \stilde_0 \right]\Prob\left[s_t=j|C,a_t, \stilde_t, \cdots, \stilde_0 \right]}{\Prob\left[\stilde_{t+1}, a_t, \stilde_t, \cdots, \stilde_0\right]}p_{t}(C)\\
            &= \frac{\sum_{i}C_{i, \stilde_{t+1}}\sum_{j}P(a_t)_{ji}b_{t,j}}{\Prob\left[\stilde_{t+1}, a_t, \stilde_t, \cdots, \stilde_0\right]}p_{t}(C)= \frac{b_t^\top P(a_t) C_{\stilde_{t+1}}}{\int_{C'} b_t^\top P C'_{\stilde_{t+1}}p_t(C')\mathrm{d}C'}p_{t}(C).
        \end{align*}
        \begin{align*}
            b_{t+1,i} &= \Prob\left[s_{t+1}=i|\stilde_{t+1},a_{t},\cdots, a_1,\stilde_0\right] \\
            &= \int_C\Prob\left[s_{t+1}=i|\stilde_{t+1},a_{t},\cdots, a_1,\stilde_0,C\right]p_t(C)\mathrm{d}C \\
            &= \int_C\frac{\Prob\left[\stilde_{t+1}|s_{t+1}=i,a_{t},\cdots, a_1,\stilde_0\right]\Prob\left[s_{t+1}=i|a_{t},\stilde_t,\cdots, a_1,\stilde_0\right]}{\Prob\left[\stilde_{t+1},a_{t},\cdots, a_1,\stilde_0\right]}p_t(C)\mathrm{d}C \\
            &= \int_C\frac{C_{i, \stilde_{t+1}}\sum_j \Prob\left[s_{t+1}=i|s_t=j,a_{t},\stilde_t,\cdots, a_1,\stilde_0\right]\Prob\left[s_{t}=j|\stilde_t,\cdots, a_1,\stilde_0\right]}{\Prob\left[\stilde_{t+1},a_{t},\cdots, a_1,\stilde_0\right]}p_t(C)\mathrm{d}C \\
            &= \int_C\frac{C_{i, \stilde_{t+1}}\sum_j P(a_t)_{ji}b_{t,j}}{\Prob\left[\stilde_{t+1},a_{t},\cdots, a_1,\stilde_0\right]}p_t(C)\mathrm{d}C = \int_C\frac{C_{i, \stilde_{t+1}}\sum_{j} P(a_t)_{ji}b_{t,j}}{\sum_{k}C_{k, \stilde_{t+1}}\sum_{j} P(a_t)_{jk}b_{t,j}}p_{t+1}(C).
        \end{align*}

        \section{Simulations}\label{sec:sim}

\begin{figure*}[h!]
            \centering
            \includegraphics[width=.24\textwidth]{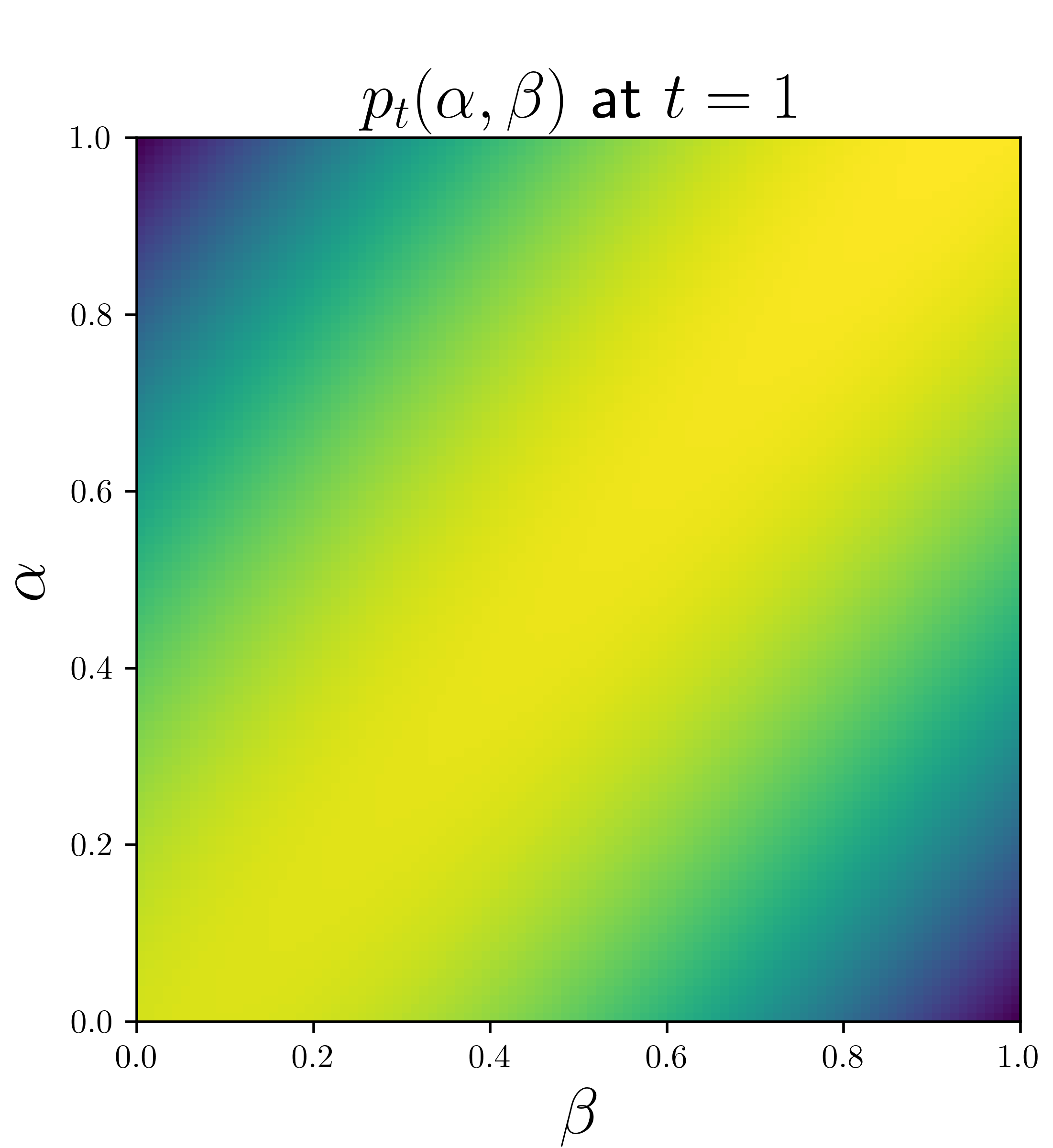}
            \includegraphics[width=.24\textwidth]{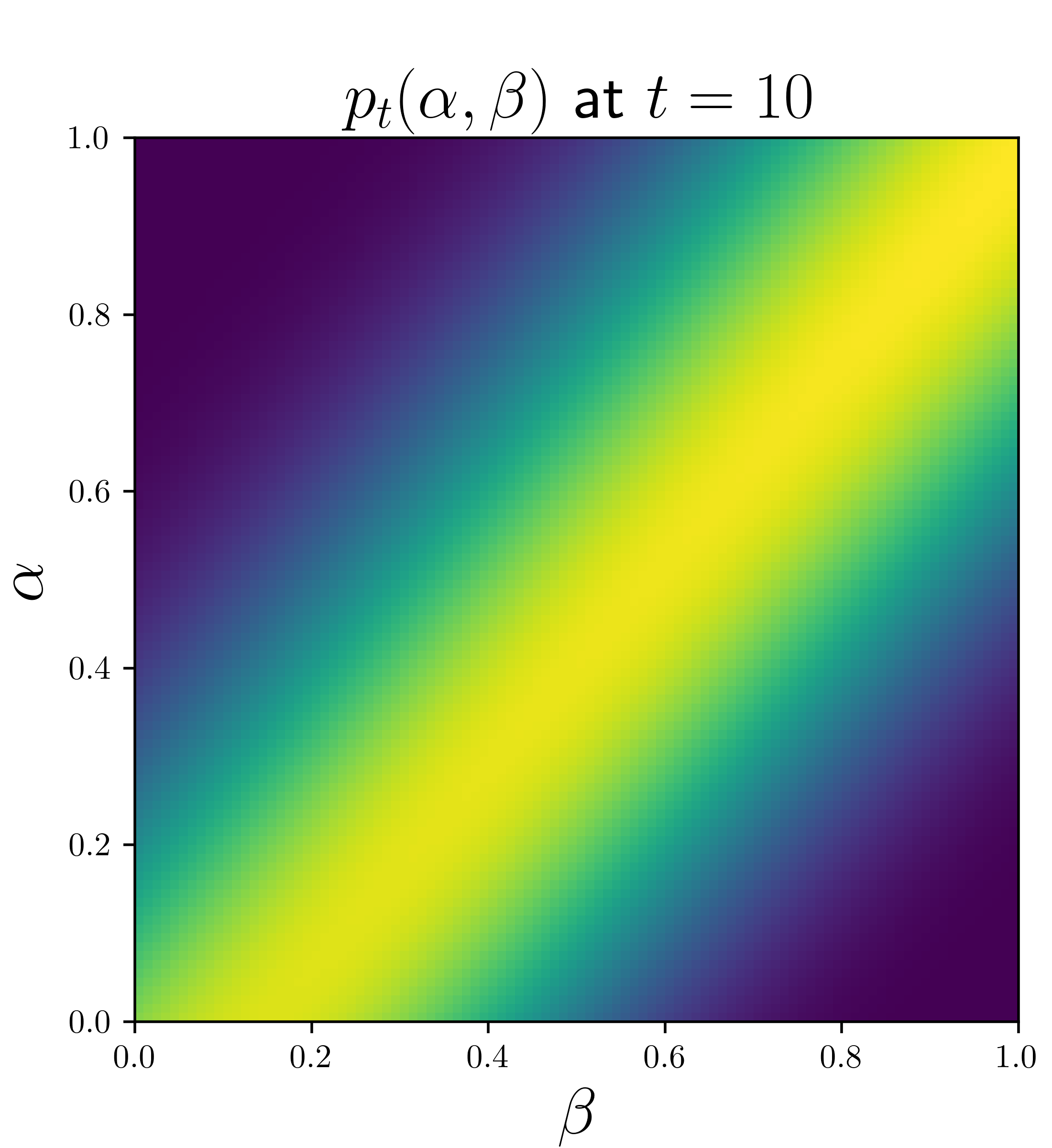}
            \includegraphics[width=.24\textwidth]{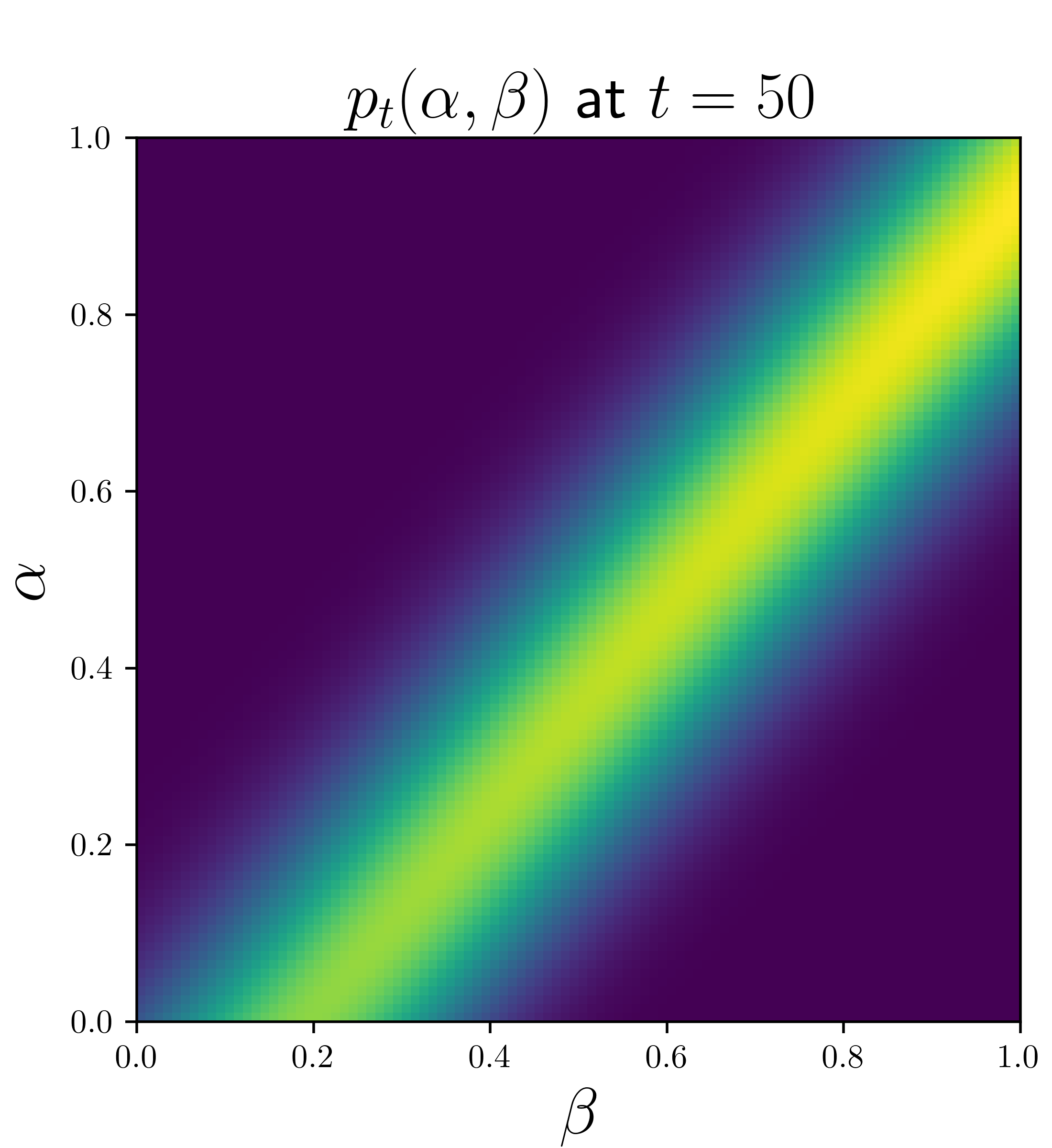}
            \includegraphics[width=.24\textwidth]{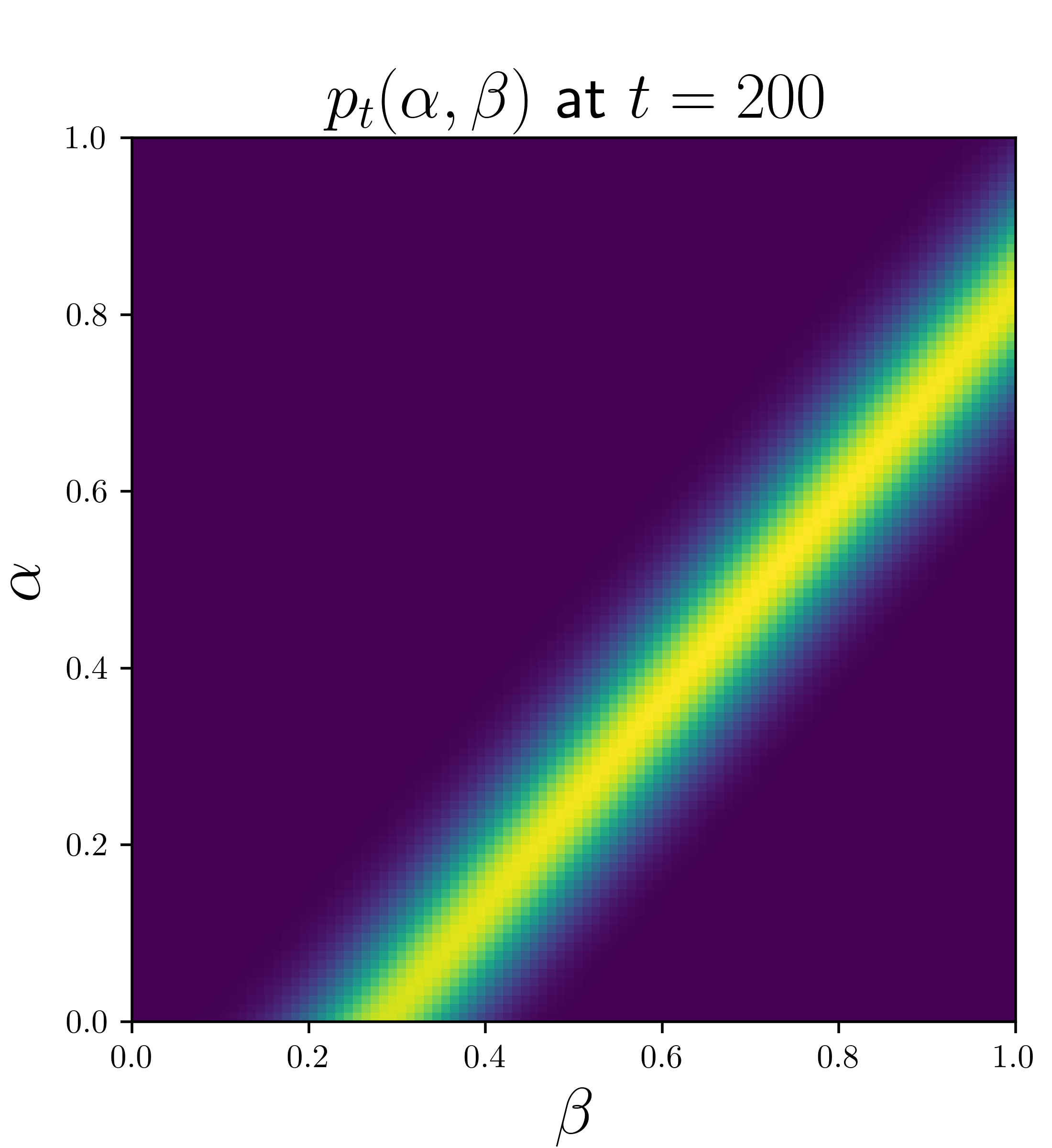}
            \caption{\(p_t(\alpha, \beta)\) for first-order Bayesian method and common stationary distributions}
            \label{fig:b1_not_unique}

            \includegraphics[width=.24\textwidth]{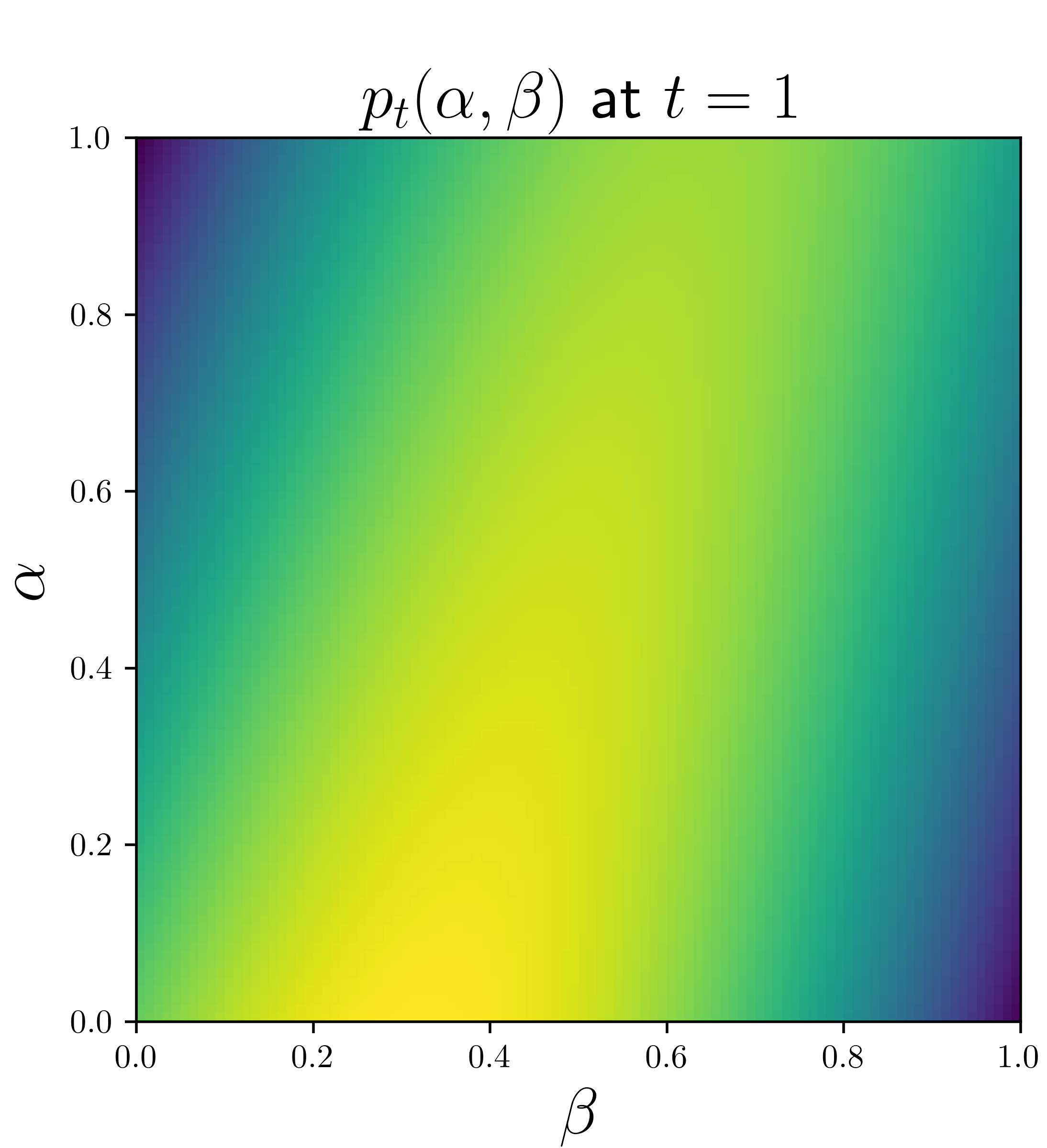}
            \includegraphics[width=.24\textwidth]{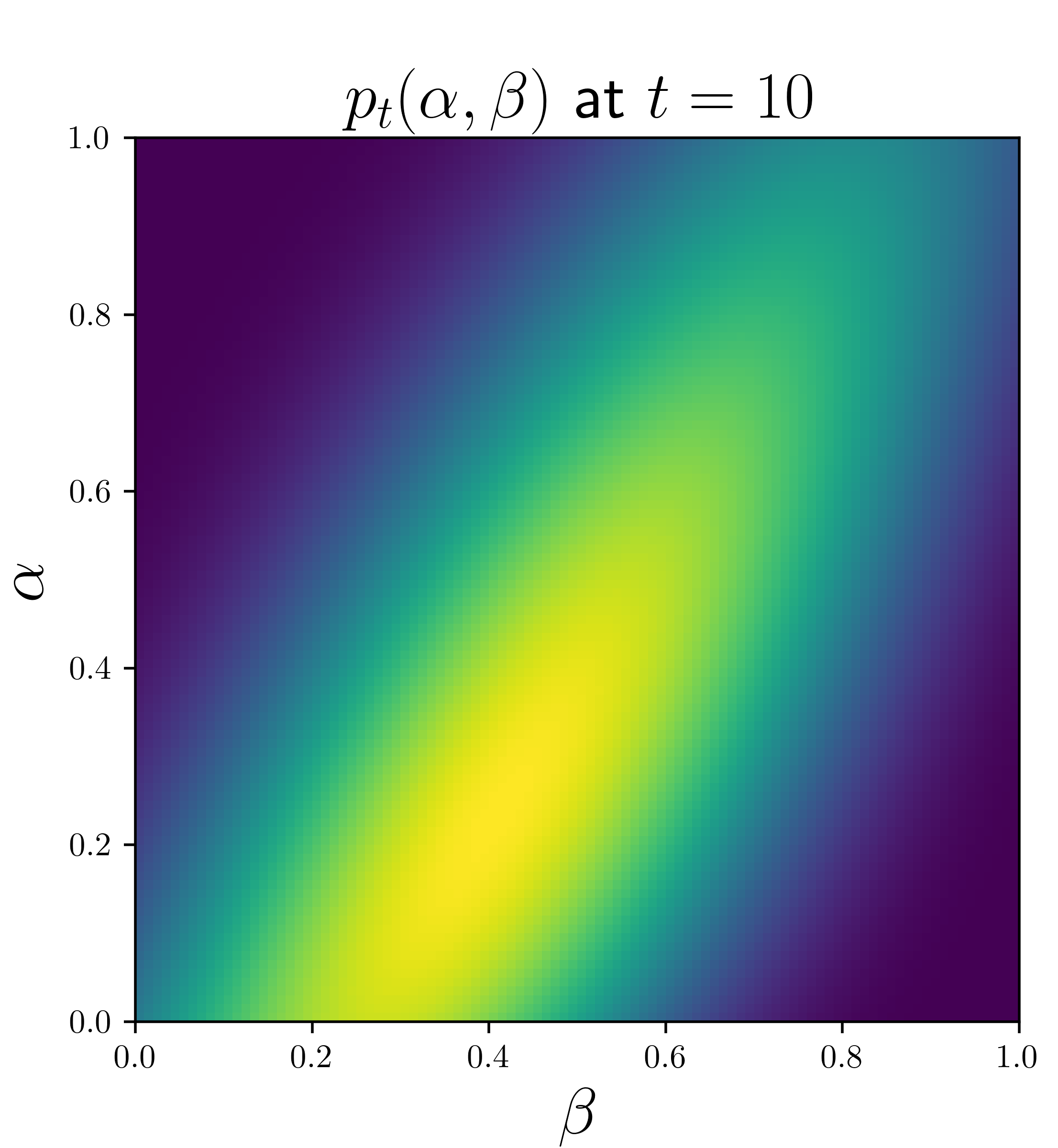}
            \includegraphics[width=.24\textwidth]{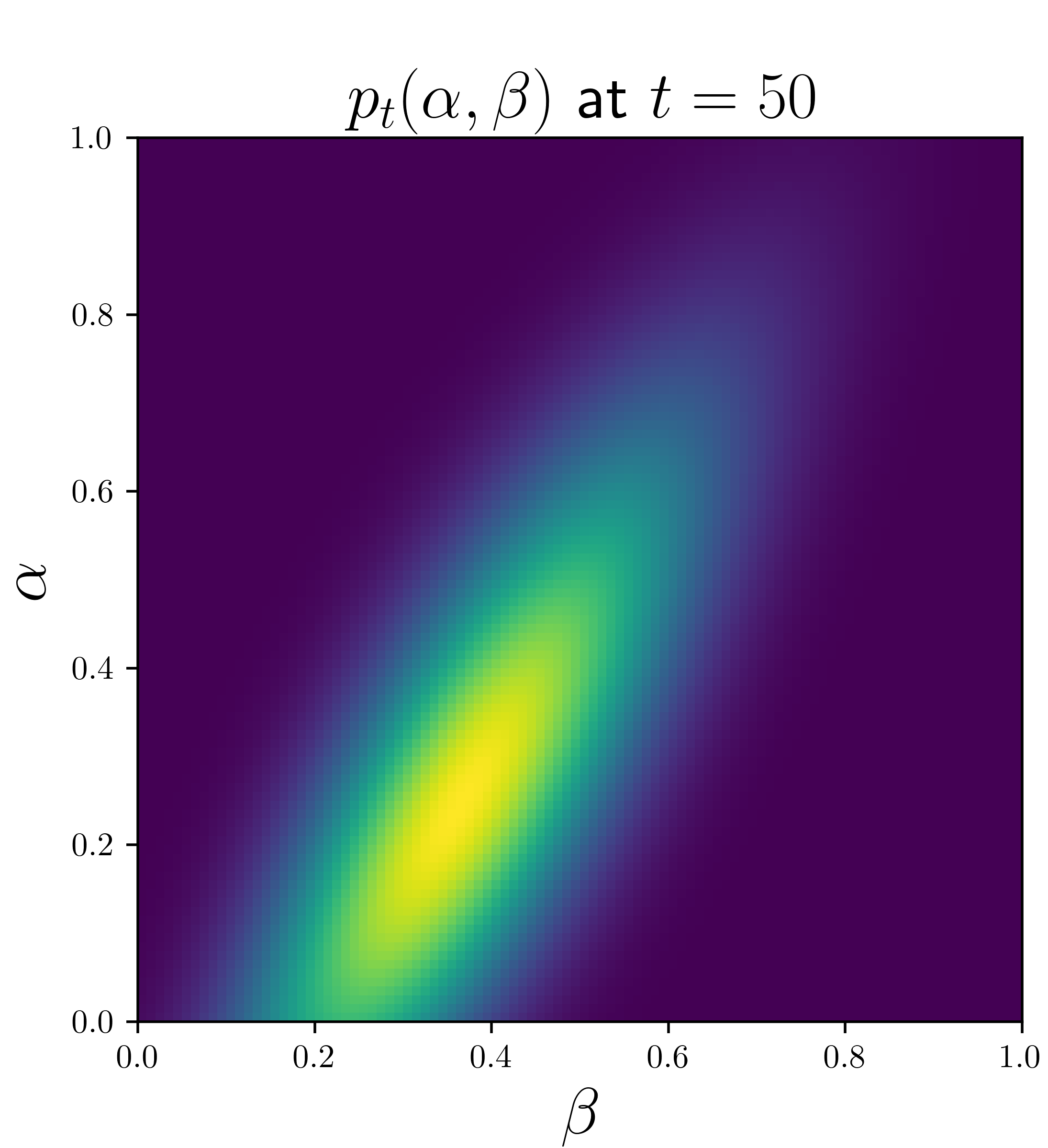}
            \includegraphics[width=.24\textwidth]{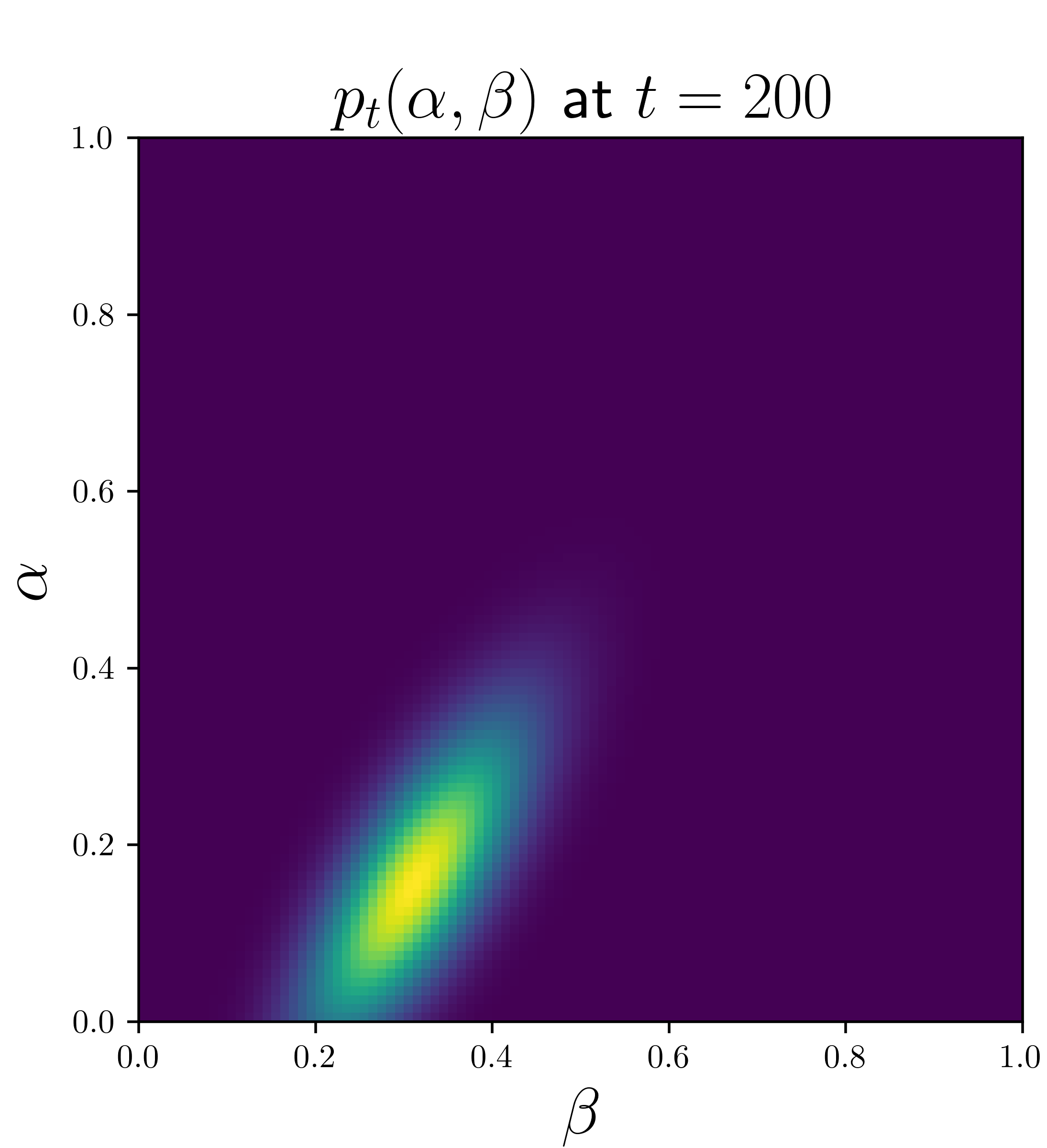}
          \caption{\(p_t(\alpha, \beta)\) for first-order Bayesian method and different stationary distributions}
          \label{fig:b1_unique}

            \includegraphics[width=.24\textwidth]{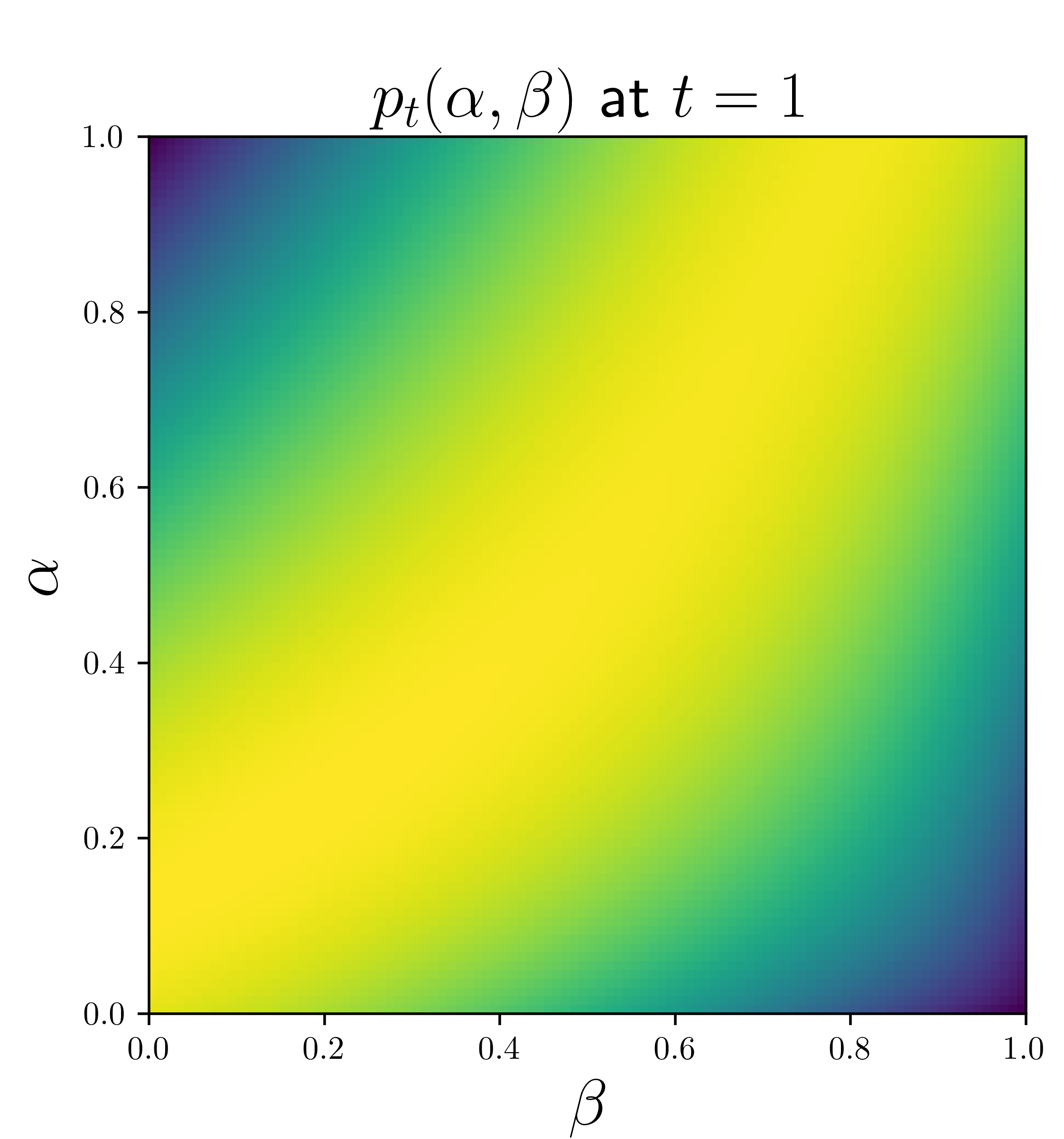}
            \includegraphics[width=.24\textwidth]{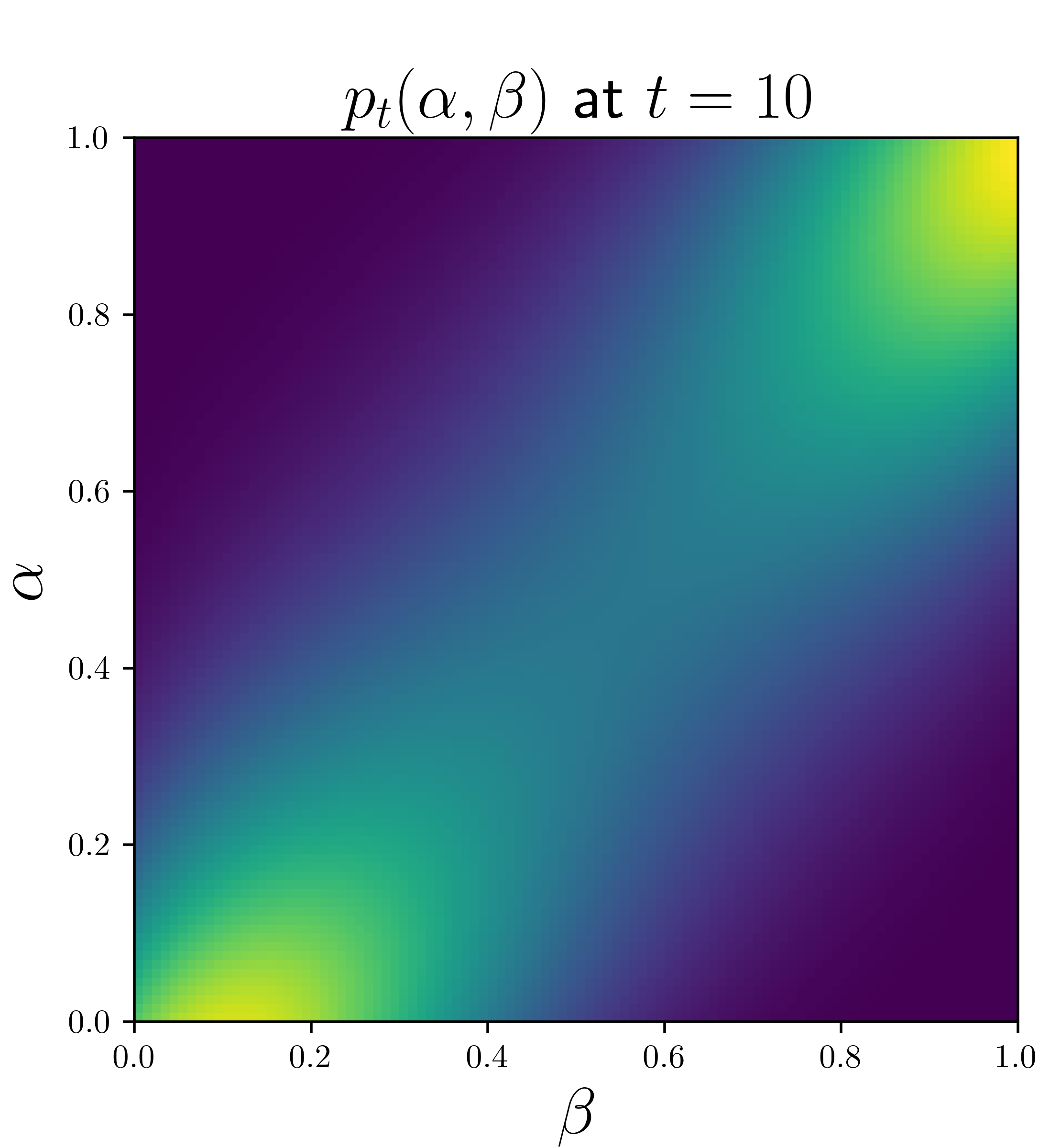}
            \includegraphics[width=.24\textwidth]{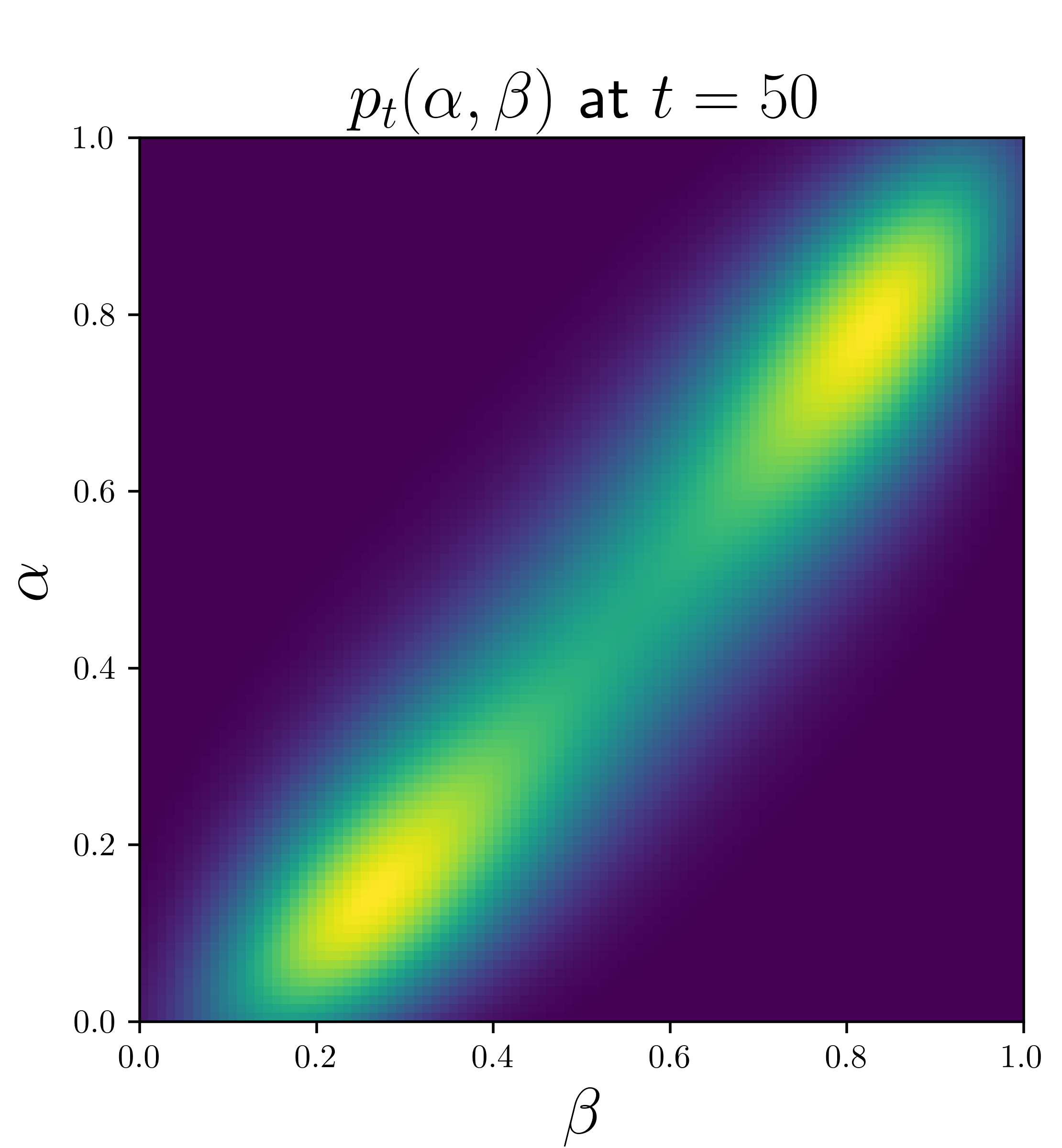}
            \includegraphics[width=.24\textwidth]{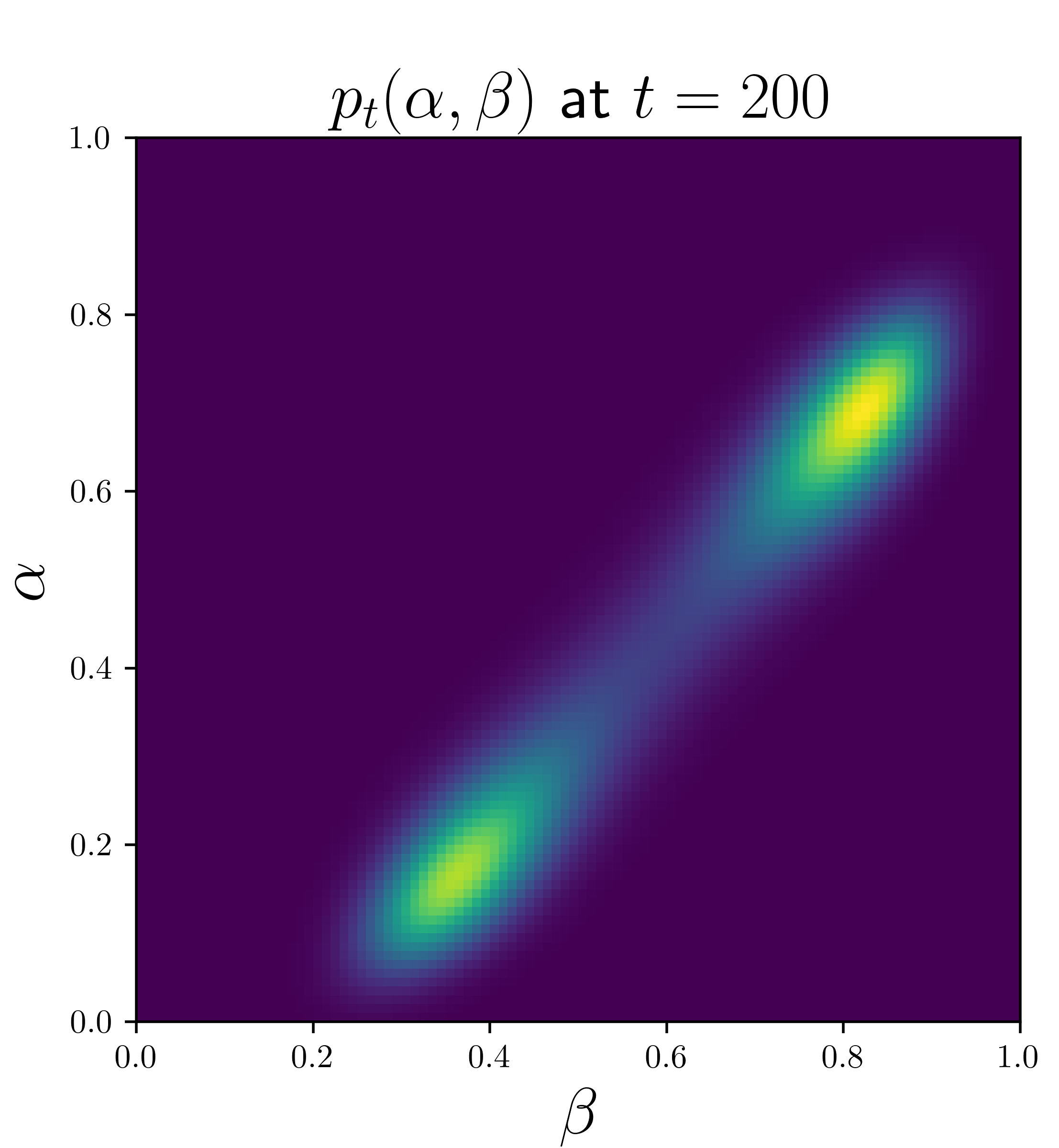}
            \caption{\(p_t(\alpha, \beta)\) for second-order Bayesian method and common stationary distributions}
            \label{fig:b2_not_unique}

            \includegraphics[width=.24\textwidth]{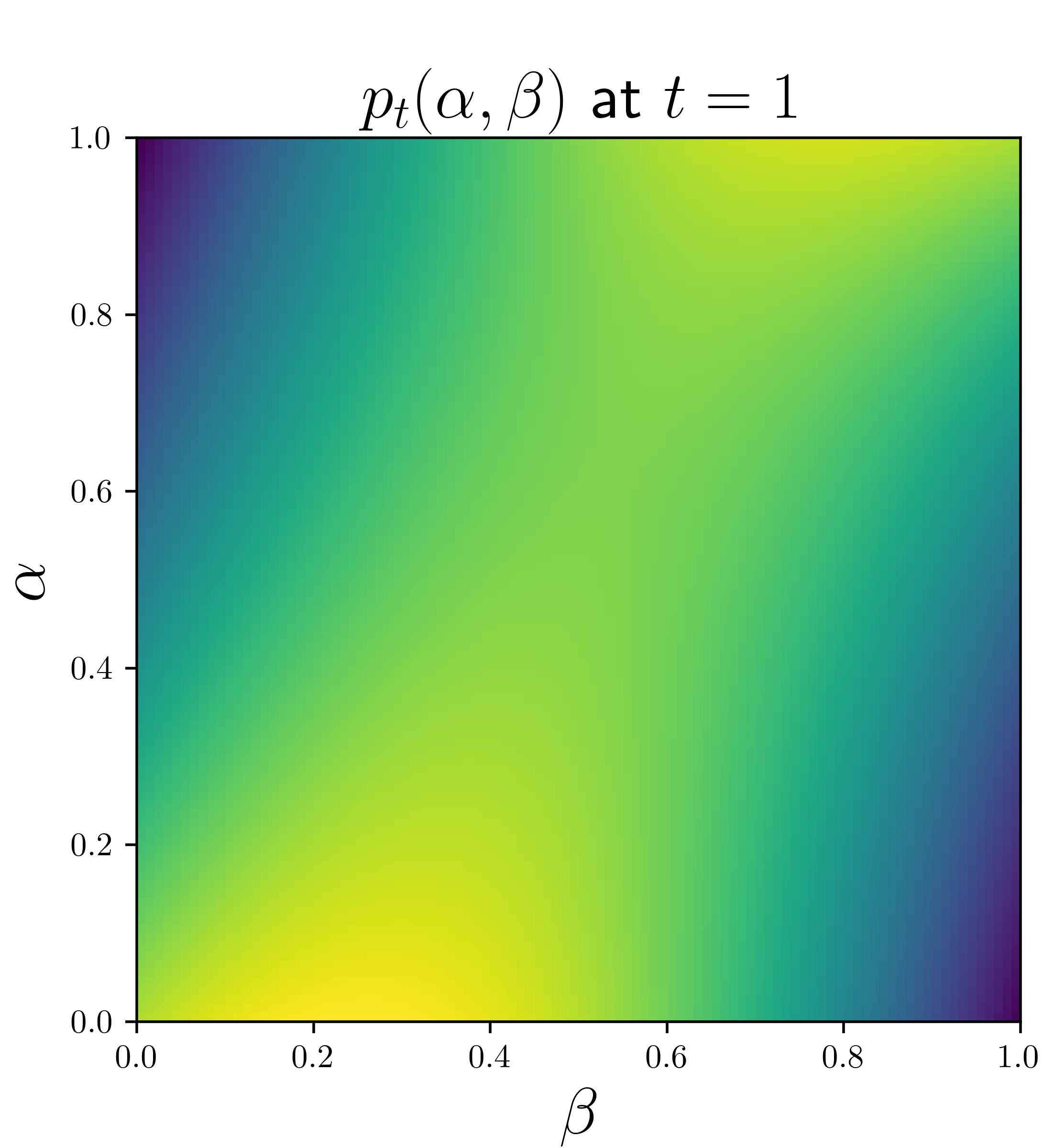}
            \includegraphics[width=.24\textwidth]{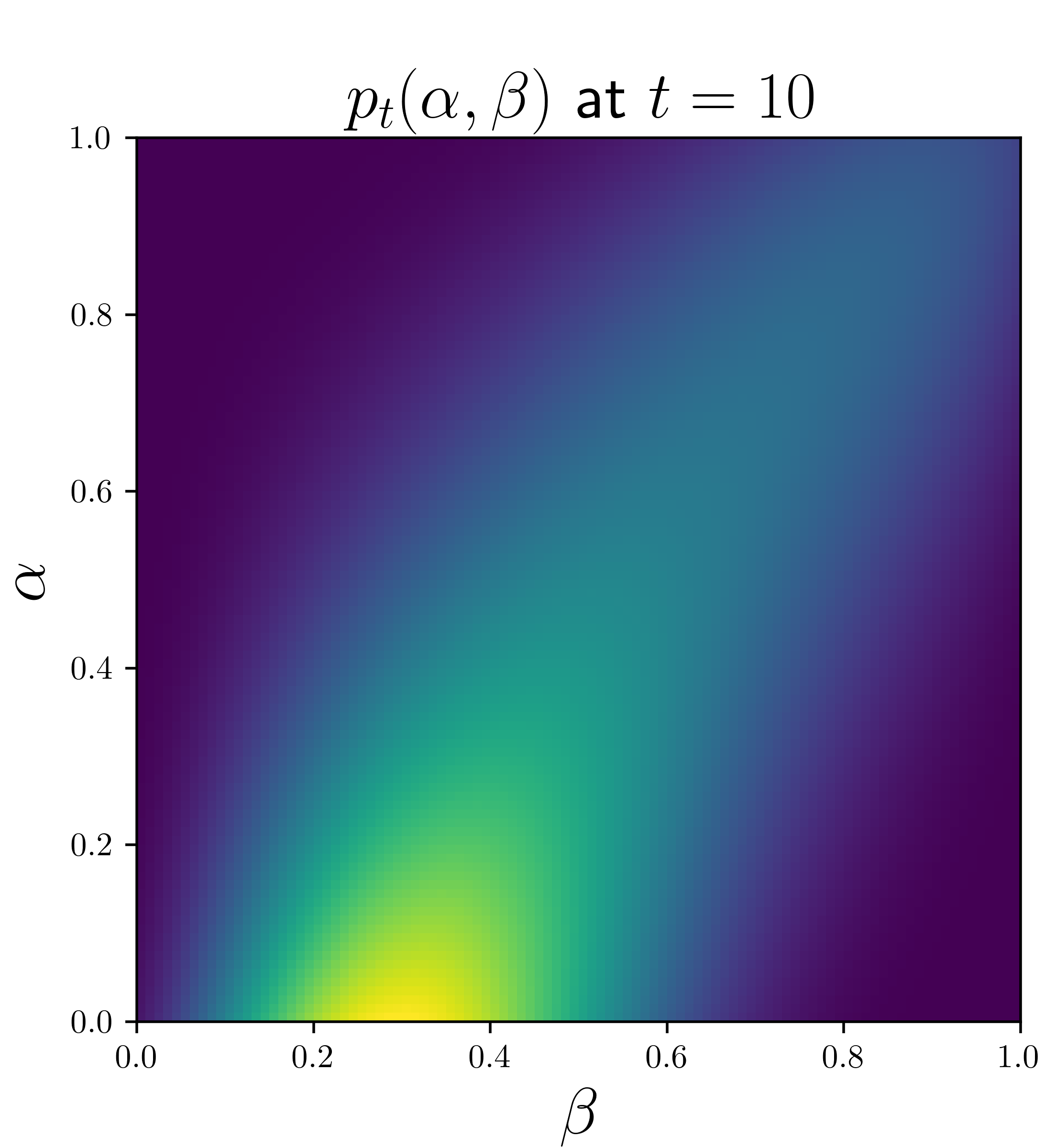}
            \includegraphics[width=.24\textwidth]{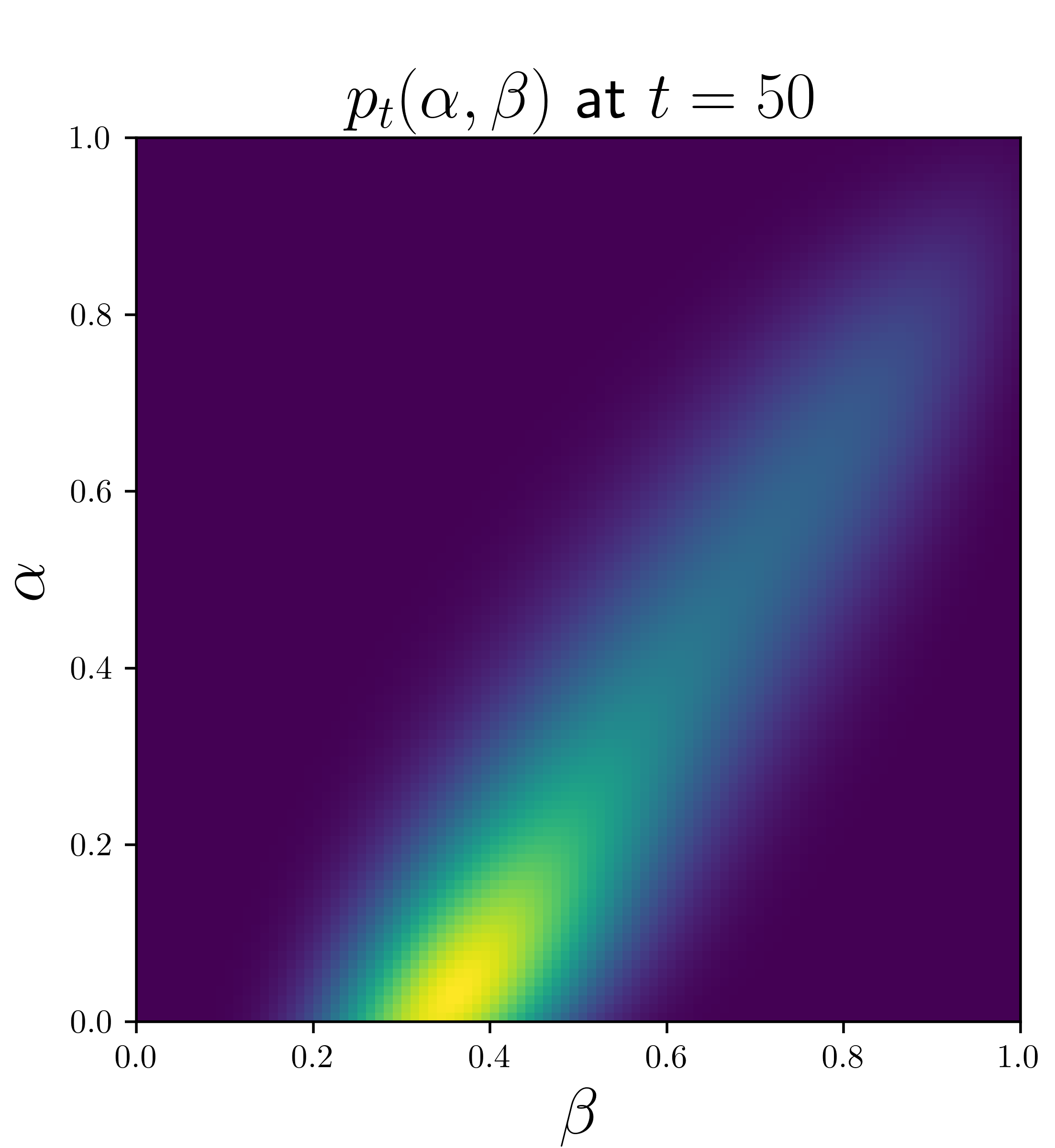}
            \includegraphics[width=.24\textwidth]{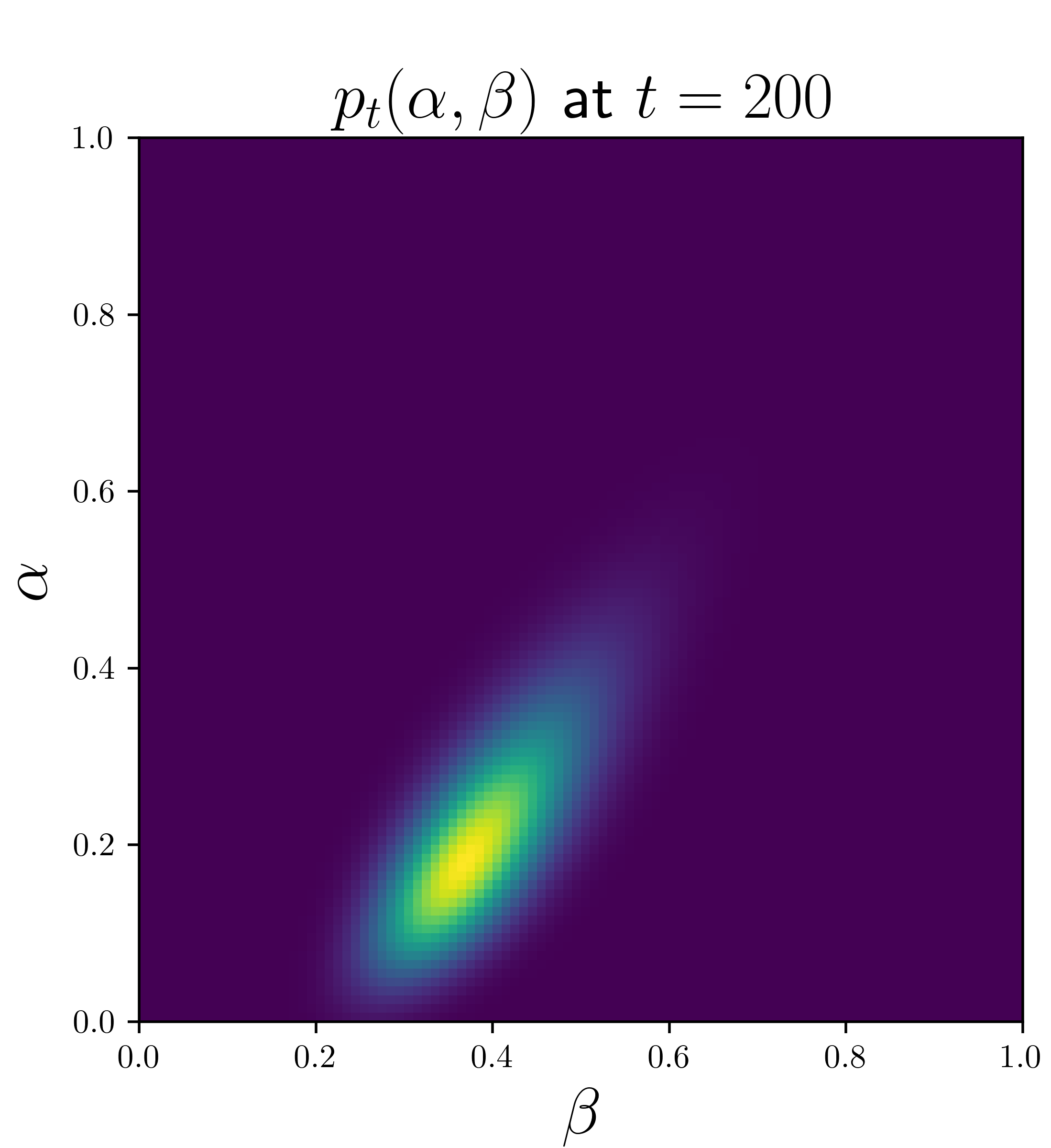}
          \caption{\(p_t(\alpha, \beta)\) for second-order Bayesian method and different stationary distributions}
          \label{fig:b2_unique}
          \label{fig01}
    \end{figure*}

In this section, we present simulations to demonstrate the effectiveness of the Bayesian approach under various settings. We choose \(n=2\) for visualization purposes. The figures \ref{fig:b1_not_unique}, \ref{fig:b1_unique}, \ref{fig:b2_not_unique}, and \ref{fig:b2_unique} depict the probability density at
\[C=\begin{bmatrix}
    1-\alpha & \alpha \\
    \beta & 1-\beta
\end{bmatrix}\]
by plotting \((\alpha, \beta)\) at different time steps, with the true values generating the observations set at $\alpha=0.4$ and $\beta=0.2$. 

The figures illustrate the performance of the first-order and second-order Bayesian methods in two scenarios: one where the transition probability matrices for different actions have a common stationary distribution, and another where they have distinct stationary distributions.

    The results of our simulations support the theoretical insights from Theorems \ref{thm:2_unique} and \ref{thm:n_unique}. Specifically, we observe convergence to the true values \((\alpha,\beta)=(0.4, 0.2)\) in scenarios with distinct stationary distributions. This is in line with the theoretical predictions for scenarios where unique solutions are expected. In contrast, in environments with common stationary distributions, we notice that neither the first-order nor the second-order Bayesian methods can uniquely pinpoint the correct solution. This outcome aligns with the theoretical anticipation of non-uniqueness in such settings. Notably, the first-order method tends to converge to a line in the \(\alpha\)-\(\beta\) plane, whereas the second-order method exhibits convergence towards two potential solutions. This empirical observation is particularly interesting as it is the same as the closed-form solution for the second possibility outlined in \eqref{eq:2_solutions} from the proof of Theorem \ref{thm:2_unique}. This outcome is visually evident in the convergence area depicted in Figure \ref{fig:b2_not_unique}, showcasing the practical implications and validity of our theoretical findings.

    \section{Conclusion and Future Work}
    This paper has tackled a particular class of the significant challenge of noisy state observations in Markov Decision Processes (MDPs) by employing a confusion matrix to model this uncertainty. We introduced two novel algorithms for noise estimation. The first, the method of second-order repetitive actions, efficiently estimates the confusion matrix \(C\) in a finite time frame. Moreover, our analysis reveals that system identifiability is contingent upon the dissimilarity of stationary distributions in action transition probabilities. Our second approach, grounded in Bayesian methods, focuses on maintaining and updating probability distributions over unknown parameters and states. This method provides the flexibility to choose actions freely, yet it faces practical computational challenges, particularly in maintaining distributions over large \(n\times n\) matrices.

    Looking to the future, several avenues for research emerge from our findings. One intriguing direction is extending our approach of second-order observation recording to higher orders, examining its impact on system identifiability and efficiency. Further theoretical exploration is needed to address the general problem of system identifiability. Particularly, A key theoretical question arises from our observation that non-identifiable systems lead to limited solution spaces: the first-order Bayesian method reduces options to a line, whereas the second-order method identifies two distinct solutions. Future research could explore whether more advanced methods exist that can fully resolve this issue, or if it can be theoretically proven that no algorithm can surpass the efficacy of the second-order method. This could significantly advance our understanding of system identifiability in MDPs with noisy observations. In addition, there is substantial scope for application-oriented research. Applying our methodologies to real-world data, exploring other classes of noisy observations within our framework, and tailoring our approaches to specific challenges in different applications offer practical avenues for future exploration. 
		

\bibliographystyle{unsrt}  
\bibliography{references}

\end{document}